\pdfoutput=1

\documentclass[11pt]{article}

\usepackage{emnlp2021}

\usepackage{times}
\usepackage{latexsym}

\usepackage[T1]{fontenc}

\usepackage[utf8]{inputenc}

\usepackage{microtype}

\usepackage{standalone}
\usepackage{tikz}
\usepackage{pgfplots}
\pgfplotsset{compat=1.14}
\usepgfplotslibrary{statistics}
\usepackage{booktabs}
\usepackage{tabularx}
\usepackage{makecell}

\def\blue#1{\textcolor{blue}{#1}}
\def\red#1{\textcolor{red}{#1}}
\def\green#1{\textcolor{olive}{#1}}
\def\violet#1{\textcolor{violet}{#1}}

\usepackage[normalem]{ulem} 

\title{Sequence Length is a Domain: \\
    Length-based Overfitting in Transformer Models}

\author{Du\v{s}an Vari\v{s} and Ond\v{r}ej Bojar \\
    Faculty of Mathematics and Physics, Charles University, \\
    Malostransk\'{e} n\'{a}m\v{e}st\'{i} 25, \\
    118 00 Prague, Czechia \\
    \texttt{\{varis,bojar\}@ufal.mff.cuni.cz}
}

\begin{document}
\maketitle
\begin{abstract}

Transformer-based sequence-to-sequence architectures, while achieving state-of-the-art results on a large number of NLP tasks, can still suffer from overfitting during training.
In practice, this is usually countered either by applying regularization methods (e.g. dropout, L2-regularization) or by providing huge amounts of training data.
Additionally, Transformer and other architectures are known to struggle when generating very long sequences.
For example, in machine translation, the neural-based systems perform worse on very long sequences when compared to the preceding phrase-based translation approaches \citep{koehn2017challenges}.

We present results which suggest that the issue might also be in the mismatch between the length distributions of the training and validation data combined with the aforementioned tendency of the neural networks to overfit to the training data.
We demonstrate on a simple string editing task and a machine translation task that the Transformer model performance drops significantly when facing sequences of length diverging from the length distribution in the training data.
Additionally, we show that the observed drop in performance is due to the hypothesis length corresponding to the lengths seen by the model during training rather than the length of the input sequence.




\end{abstract}

\section{Introduction}

Current state-of-the-art Transformer-based sequence generation models, either fine-tuned for chosen downstream tasks \citep{devlin2019bert}, or trained from scratch for specific tasks such as machine translation \citep{vaswani2017attention} or speech recognition \citep{nguyen2019interspeech}, more and more often achieve performance comparable to that of humans \citep{hassan2018parity, popel2020cubbitt, nguyen2020superhuman}.
However, such models frequently require billions of trainable parameters together with huge amounts of data (billions of tokens) to reach such performance \citep{brown2020gpt3}.

The good performance on held-out test sets seems to confirm the good generalization power of these models, although the inherent strong biases, sometimes leading to the use of a foul and toxic language, preserving stereotypes, etc., are well acknowledged \citep{gehman2020toxicity}. \citet{brown2020gpt3} claim that their Transformer model is also capable of simple arithmetics, however, it is yet to be validated whether the model truly learns the arithmetic algorithms or simply encodes a lookup table for a subset of specific examples.

In this paper, we argue that the assumed generalization power of the current state-of-the-art Transformer-based language generators does not come from the architecture itself but rather from the sheer volume of training data and the model's ability to exploit the similarities between the training and validation data.
We demonstrate how the Transformer-based sequence-to-sequence models fail when the target sequence lengths of the training and validation data do not match.
We show that this holds not only for very long test sequences but can be observed even with short sequences if they are omitted from the training data.
Furthermore, we show that we can artificially improve the test performance on longer sequences by only using shorter training sequences and concatenating them into longer training examples.


We do not argue about Transformer's (in)ability to handle long-distance dependencies, but our results suggest that a considerably simpler reason of mismatching sequence length can also contribute to the performance drop.
We think that our findings can lead to better understanding of the Transformer architecture and help to design better training schemes (e.g. curriculum learning).






\section{Related Work}


The problem of modeling very long sequences has been studied mainly in the context of recurrent neural networks (RNNs).
Early studies showed that using LSTMs \citep{sutskever2014seq2seq} and introducing attention \citep{bahdanau2015neural, luong2015attention} can improve the model performance on long sequences.
However, these models still got outperformed on long sequences by phrase-based models
 \citep{koehn2017challenges}.
This problem was not resolved with the introduction of Transformers \citep{vaswani2017attention}.
Surprisingly, even though there were previous studies explaining the weaknesses of RNNs with respect to long sequence modeling \citep{hochreiter1997lstm, hochreiter1998vanishing}, similar analyses are yet to be done for Transformers which are fundamentally different from RNNs.

There is an ongoing debate about the proper way of splitting the available data to training and evaluation subsets.
\citet{gorman2019standard-splits} show that using only standard dataset splits can lead to a biased evaluation resulting in overestimating the generalization ability of the model.
Furthermore, \citet{sogaard2020random-splits} argue that even using randomly sampled dataset splits does not solve the overestimation problem.
They instead suggest using multiple test sets possibly of an adversarial nature to properly evaluate the generalization ability of the model.

In the following experiments, we evaluate vanilla Transformer on such adversarial splits created with respect to the lengths of the modeled sequences.
Although similar analyses were performed in the past \citep{neishi2019relation,kondo2021sentence}, it was at a smaller scale and mainly in the context source-side length bucketing.


\section{Experiments}

We demonstrate the lack of ability to generalize to sequences of lengths not seen during training on two separate tasks: \textit{string editing} and \textit{machine translation} (MT).

We use Fairseq framework for sequence-to-sequence learning \citep{ott2019fairseq} in our experiments.\footnote{\url{https://github.com/pytorch/fairseq}}
Details about the model parameters and training are available in Appendix~\ref{app:model}.


\subsection{String Editing Operations}

\begin{table}[t]
\begin{center}
\small
\begin{tabular}{l|l}
\toprule
Input & Output \\
\midrule
push 1 | 1 0 1 0 & 1 0 1 0 1\\
reverse $-$ | 1 0 0 1 1 & 1 1 0 0 1 \\
\bottomrule
\end{tabular}
\caption{Input and output example for \texttt{push} and \texttt{reverse} tasks. Hyphen ($-$) indicates an empty argument for the latter task.}
\label{tab:string-example}
\end{center}
\end{table}

\begin{table}[t]
\begin{center}
\small
\begin{tabular}{r|rrr}
\toprule
 & 0-10 & 11-15 & 16-20 \\
\midrule
copy & 62.6 & 100.0 & 0.0 \\
push & 59.1 & 100.0 & 0.0 \\
pop & 0.1 & 100.0 & 0.0 \\
shift & 52.5 & 100.0 & 0.0 \\
unshift & 41.2 & 100.0 & 0.0 \\
reverse & 0.0 & 84.4 & 0.0 \\
\midrule
all & 15.822 & 97.5 & 0.978 \\
\bottomrule
\end{tabular}
\caption{Accuracy (in $\%$) of models trained on various string editing tasks using only training data from the 11-15 length bucket evaluated on datasets with different sequence lengths. Each model was evaluated on its respective task domain.} 
\label{tab:string-buckets}
\end{center}
\end{table}

In the first set of experiments, we focus on learning simple string editing algorithms.
We chose this task because we think it is an interesting alternative to standard NLP tasks that often struggle with evaluation ambiguity (multiple possible outputs in MT or text generation with nuanced degree of quality) and proper training/validation separation
(partial overlap between train and test sentences leading to lack of clarity how much model actually generalizes to new inputs).


We chose to study the following tasks:
\begin{itemize}
    \item \textit{copy}: copy the input sequence to the output,
    \item \textit{unshift X}, \textit{push X}: add a single character (X) to the beginning or the end of the sequence respectively
    \item \textit{shift}, \textit{pop}: remove a single character from the beginning or the end respectively,
    \item \textit{reverse}: reverse the character order in the input sequence
\end{itemize}

\begin{table*}[ht!]
\begin{center}
\small
\begin{tabular}{r|rrrrrrrr}
\toprule
Bucket & 0-10 & 11-20 & 21-30 & 31-40 & 41-50 & 51-60 & 61-70 & 71-80 \\
\midrule
\# of sent. pairs (M) & 30.9 & 18.0 & 7.5 & 3.9 & 2.3 & 1.2 & 0.7 & 0.4 \\
\# of tokens (M) & 375.3 & 502.6 & 361.6 & 268.9 & 198.9 & 132.6 & 87.3 & 59.5 \\
\bottomrule
\end{tabular}
\caption{Sizes of the respective training buckets (created based on the target sequence length) in millions of sentence pairs and millions of tokens (after tokenization and applying BPE, combined source and target size).}
\label{tab:dataset-sizes}
\end{center}
\end{table*}

As for the experiment setup, we generate a dataset of sequences consisting of two characters (e.g. $0$ and $1$), separated by whitespace, with no duplicate sequences.
Then, we split the dataset into three separate buckets according to sequence lengths, $0-10$, $11-15$ and $16-20$ respectively. We sample 1,000 sequences from the $0-10$ and $16-20$ buckets for test-time evaluation.
We split the $11-15$ bucket into a validation (1,000 examples), test (1,000 examples) and training (28,000 examples) from a sample of 30k examples without repetition.

Given these data splits, we create datasets for each task by adding the task label, character ($0$, $1$ for unshift and push, $-$ for others) and a separator ($|$) to the beginning of each sequence.\footnote{The arguments for unshift and push are sampled from a Bernoulli distribution with $0$ character having $p=0.5$.}
We create target sequences for each task according to the respective task definition.
Table~\ref{tab:string-example} shows examples of the networks inputs.

\begin{figure*}[t]
    \centering
    \begin{minipage}[b]{\linewidth}
        \centering
        \def\verticalstep{0.8}
\small
\pgfplotscreateplotcyclelist{buckets}{
semithick,blue,mark=o\\%
semithick,red,mark=o\\%
semithick,green,mark=o\\%
semithick,orange,mark=o\\%
semithick,cyan,mark=o\\%
semithick,blue,mark=triangle*\\%
semithick,red,mark=triangle*\\%
semithick,green,mark=triangle*\\%
semithick,orange,mark=triangle*\\%
semithick,cyan,mark=triangle*\\%
}

\begin{tikzpicture}[]
  \begin{axis}[cycle list name=buckets, 
               xlabel={},
               ylabel={BLEU},
               xmax=120,
               width=0.8\textwidth,
               height=0.27\textwidth,
               x label style={at={(axis description cs:0.5,-0.1)},anchor=north},
               y label style={at={(axis description cs:-0.1,.5)},rotate=0,anchor=south},
               legend style={nodes={scale=0.5, transform shape},at={(1,1)},anchor=north east}]

    \addplot+ table [x=BUCKET, y=BLEU, col sep=comma] {./data/buckets/buckets.10.csv};
    \addplot+ table [x=BUCKET, y=BLEU, col sep=comma] {./data/buckets/buckets.20.csv};
    \addplot+ table [x=BUCKET, y=BLEU, col sep=comma] {./data/buckets/buckets.30.csv};
    \addplot+ table [x=BUCKET, y=BLEU, col sep=comma] {./data/buckets/buckets.40.csv};
    \addplot+ table [x=BUCKET, y=BLEU, col sep=comma] {./data/buckets/buckets.50.csv};
    \addplot+ table [x=BUCKET, y=BLEU, col sep=comma] {./data/buckets/buckets.60.csv};
    \addplot+ table [x=BUCKET, y=BLEU, col sep=comma] {./data/buckets/buckets.70.csv};
    \addplot+ table [x=BUCKET, y=BLEU, col sep=comma] {./data/buckets/buckets.80.csv};
    \addplot+ table [x=BUCKET, y=BLEU, col sep=comma] {./data/buckets/buckets.csv};
    \addlegendentry{TrainBucket = 10}
    \addlegendentry{TrainBucket = 20}
    \addlegendentry{TrainBucket = 30}
    \addlegendentry{TrainBucket = 40}
    \addlegendentry{TrainBucket = 50}
    \addlegendentry{TrainBucket = 60}
    \addlegendentry{TrainBucket = 70}
    \addlegendentry{TrainBucket = 80}
    \addlegendentry{Full CzEng}
  \end{axis}
\end{tikzpicture}
        \def\verticalstep{0.8}
\small
\pgfplotscreateplotcyclelist{buckets}{
semithick,blue,mark=o\\%
semithick,red,mark=o\\%
semithick,green,mark=o\\%
semithick,orange,mark=o\\%
semithick,cyan,mark=o\\%
semithick,blue,mark=triangle*\\%
semithick,red,mark=triangle*\\%
semithick,green,mark=triangle*\\%
semithick,orange,mark=triangle*\\%
semithick,cyan,mark=triangle*\\%
}

\begin{tikzpicture}[]
  \begin{axis}[cycle list name=buckets, 
               xlabel={Test Bucket},
               ylabel={Hyp/Ref Ratio},
               xmax=120,
               width=0.8\textwidth,
               height=0.26\textwidth,
               x label style={at={(axis description cs:0.5,-0.1)},anchor=north},
               y label style={at={(axis description cs:-0.1,.5)},rotate=0,anchor=south},
               legend style={nodes={scale=0.5, transform shape},at={(1,1)},anchor=north east}]

    \addplot+ table [x=BUCKET, y=RATIO, col sep=comma] {./data/ratios/ratios.10.csv};
    \addplot+ table [x=BUCKET, y=RATIO, col sep=comma] {./data/ratios/ratios.20.csv};
    \addplot+ table [x=BUCKET, y=RATIO, col sep=comma] {./data/ratios/ratios.30.csv};
    \addplot+ table [x=BUCKET, y=RATIO, col sep=comma] {./data/ratios/ratios.40.csv};
    \addplot+ table [x=BUCKET, y=RATIO, col sep=comma] {./data/ratios/ratios.50.csv};
    \addplot+ table [x=BUCKET, y=RATIO, col sep=comma] {./data/ratios/ratios.60.csv};
    \addplot+ table [x=BUCKET, y=RATIO, col sep=comma] {./data/ratios/ratios.70.csv};
    \addplot+ table [x=BUCKET, y=RATIO, col sep=comma] {./data/ratios/ratios.80.csv};
    \addplot+ table [x=BUCKET, y=RATIO, col sep=comma] {./data/ratios/ratios.csv};
    
    \coordinate (A) at (0,1.0);
    \coordinate (O1) at (rel axis cs:0,0);
    \coordinate (O2) at (rel axis cs:1,0);
    \draw [black,sharp plot,dashed] (A -| O1) -- (A -| O2);
    
    \addlegendentry{TrainBucket = 10}
    \addlegendentry{TrainBucket = 20}
    \addlegendentry{TrainBucket = 30}
    \addlegendentry{TrainBucket = 40}
    \addlegendentry{TrainBucket = 50}
    \addlegendentry{TrainBucket = 60}
    \addlegendentry{TrainBucket = 70}
    \addlegendentry{TrainBucket = 80}
    \addlegendentry{Full CzEng}
  \end{axis}
\end{tikzpicture}
    \end{minipage} 
    \caption{\textbf{Top}: Varying performance of Transformers on test data trained only on the data from a specific target-side length bucket (various lines) when evaluated on a specific test bucket (x-axis). When the train-test sentence length difference increases, the performance drops. Note that BLEU scores are not directly comparable across different test sets (i.e. horizontally). Within each test set, we see that the Full CzEng and the training bucket of the matching length are the two best results. \textbf{Bottom}: Average ratio between a hypothesis and reference. Dashed line indicates a ratio of 1.0. Systems trained on short sentences produce short outputs, systems trained on long sentences produce up to 10x longer outputs (Train Bucket 80 evaluated on Test Bucket 10).}
    \label{fig:mt-length}
\end{figure*}
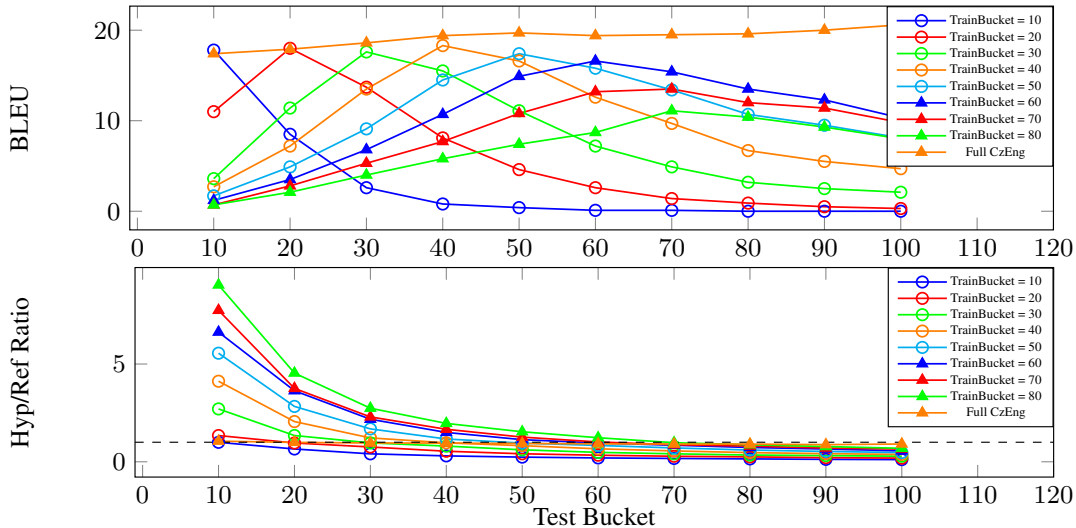

For each task, we train a separate network on the 11-15 training data. Model details are available in Appendix~\ref{app:model-string}.
We evaluate the models by measuring accuracy $ACC=num\_correct/num\_examples$, where $num\_correct$ is the number of exact matches between the hypothesis and reference strings.
Table~\ref{tab:string-buckets} shows the accuracy of the models trained on each task and evaluated on the varying test set buckets.
We can see that the models generalize very well on the unseen sequences with length similar to the training sequences, all reaching the perfect accuracy except the \textit{reverse} task.
On the other hand, when facing shorter or longer sequences, the performance drops significantly.

Table~\ref{tab:string-buckets} also shows results of the training a network on all tasks simultaneously (\textit{all}; by concatenating and shuffling respective training data and performing evaluation on the concatenation of the respective testsets).
The resulting performance is similar to that of a single-task model.

These results suggest that the length distribution similarity between the training and validation data is important and that the vanilla Transformer decoder is prone to overfitting to the sequence lengths seen during training.

\subsection{Machine Translation}

To see whether our findings within the string editing tasks also hold for natural language which has more complex structure, we perform similar experiments on English-Czech translation.

We use CzEng~2.0\footnote{\url{https://ufal.mff.cuni.cz/czeng}} \citep{kocmi2020announcing} as our training corpus, a concatenation of WMT2020 \citep{wmt2020proceedings} \texttt{newstest13-16} as held-out test set and a concatenation of \texttt{newstest17-20} for final evaluation.\footnote{We download the \texttt{newstest} corpora using SacreBLEU \citep{post2018sacrebleu}.}
We tokenize our data using Moses tokenizer.\footnote{\url{https://github.com/moses-smt/mosesdecoder.git}}
We use byte-pair encoding \citep{sennrich2016neural} on our training data, to create subword segmentation of size 30k.\footnote{\url{https://github.com/rsennrich/subword-nmt.git}}
We split all tokenized and BPE-segmented datasets into buckets of sizes 1-10, 11-20, ..., 91-100 (labeled as 10, 20, ..., 100 respectively) based on the number of tokens on the target side. Table~\ref{tab:dataset-sizes} shows the sizes of the respective training corpora.
We train a separate model for each training bucket.
Details on the model hyper-parameters are available in Appendix~\ref{app:model-mt}.

We evaluate how the length of the training data affects the performance with respect to the length of the test data using BLEU \cite{papineni2002bleu}, namely the SacreBLEU implementation \citep{post2018sacrebleu}.\footnote{\url{https://github.com/mjpost/sacrebleu}}
Figure~\ref{fig:mt-length} (Top) shows that regardless of the training bucket, the model performs best when presented with data of target-side length similar to the length of the training data.
This confirms that the model overfits to the length of the training data, affecting its performance even on shorter sentences.
The performance further decreases with increasing train-test length difference, although it needs to be noted that the BLEU scores between different testset buckets are not directly comparable due to the nature of the scoring metric and the fact that each testset bucket contains different test examples.
Figure~\ref{fig:mt-length} (Bottom) explains the main reason behind the BLEU decrease: the increased hypothesis/reference length ratio,
further supporting the length overfitting argument.
Note that the lower performance of the models trained on the 70 and 80 buckets migth be due to significantly smaller size of training data (< 1M sentence pairs).
In Appendix~\ref{app:ex}, we also provide a case study of the models trained on various length buckets.

The length-controlled experiment results presented by \citet{neishi2019relation}, while not directly focused on exploring the target-side length overfitting phenomenon, point to a similar behavior of vanilla Transformers with regards to both longer and shorter test sentences.
Based on their results, the replacement of the absolute positional embeddings with a variation of relative-position embeddings \citep{shaw2018self,neishi2019relation} seems like a promising approach towards alleviating the length overfitting problem.

To see whether we can exploit the target-side length overfitting behaviour, we also set up a separate experiment, similar to \citet{kondo2021sentence}. We take the 10, 20 and 30 training buckets and concatenate the sentences in each of them to create synthetic datasets with target-side lengths 51-60 (containing on average 6, 3 and 2 sentences per training example, respectively).
We apply the same training strategy using the synthetic data to see how strongly can the length of the training examples (although artificial) affect the model performance on the test examples of similar length.

\begin{figure}[t]
    \centering
    \def\verticalstep{0.8}

\pgfplotscreateplotcyclelist{buckets}{
semithick,blue,mark=o\\%
semithick,red,mark=o\\%
semithick,green,mark=o\\%
semithick,orange,mark=o\\%
semithick,cyan,mark=o\\%
semithick,blue,mark=triangle*\\%
semithick,red,mark=triangle*\\%
semithick,green,mark=triangle*\\%
semithick,orange,mark=triangle*\\%
semithick,cyan,mark=triangle*\\%
}
\small
\begin{tikzpicture}[]
  \begin{axis}[cycle list name=buckets, 
               xlabel={Test Bucket},
               ylabel={BLEU},
               width=0.45\textwidth,
               height=0.27\textwidth,
               xtick={0, 20, 40, 60, 80, 100},
               x label style={
                    at={(axis description cs:0.5,-0.1)},
                    anchor=north},
               y label style={
                    at={(axis description cs:-0.09,.5)},
                        rotate=0,
                        anchor=south},
               legend style={nodes={scale=0.5, transform shape},at={(1,0.4)}, anchor=north east}]

    \addplot+ table [x=BUCKET, y=BLEU, col sep=comma] {./data/buckets/buckets.concat.10.csv};
    \addplot+ table [x=BUCKET, y=BLEU, col sep=comma] {./data/buckets/buckets.concat.20.csv};
    \addplot+ table [x=BUCKET, y=BLEU, col sep=comma] {./data/buckets/buckets.concat.30.csv};
    \addplot+ table [x=BUCKET, y=BLEU, col sep=comma] {./data/buckets/buckets.60.csv};

    \addlegendentry{TrainBucket.Concat = 10}
    \addlegendentry{TrainBucket.Concat = 20}
    \addlegendentry{TrainBucket.Concat = 30}
    \addlegendentry{TrainBucket = 60}
  \end{axis}
\end{tikzpicture}
    \caption{Comparison of the performance of a model trained on genuine data from the 60-length bucket with models trained on synthetic 60-length datasets created by concatenation of 10, 20 and 30-bucket sentences respectively.}
    \label{fig:mt-concat}
\end{figure}
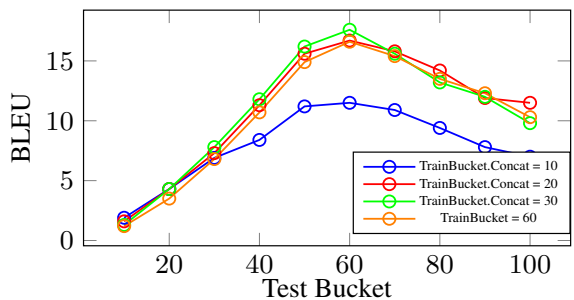

Figure~\ref{fig:mt-concat} shows that the simple concatenation of shorter training sentence pairs can lead to a performance similar to the model trained on the genuinely longer sentences. Only the performance of the model trained on  
the concatenation of very short sentences
(the line ``TrainBucket.Concat=10'' in Figure~\ref{fig:mt-concat}) drops significantly, possibly
because the model does not learn to handle any dependencies beyond the length of 10 and such dependencies seem to emerge in test sentences with length over 40, where the model starts to underperform.

\citet{kondo2021sentence} show that augmenting the existing training data with additional training examples that were created by concatenation of shorter sentences can help to improve model performance on very long sentences.
Our results show that the synthetic concatenated data on their own can be sufficient to train a model that is competitive when applied to sentences from the similar target-length domain as the training examples.
We also argue that due to a different bucket preparation strategy (based on the source-length in the previous work), the target-side length overfitting phenomenon is not as clear in \citet{kondo2021sentence} as in our work.
In Appendix~\ref{app:src-experiments}, we provide additional results from the experiments where the dataset bucketing is based on the source-side length instead of the target-side length for comparison.



\section{Conclusion}

We showed in our targeted experiment that vanilla Transformer sequence-to-sequence models have a strong tendency to overfit with regard to the target-side length of the training sequences.
On a simple algorithmic task, we documented that Transformer can generalize very well to unseen examples \emph{within the same length bucket} but falls short if the same task is required for input of a different length, shorter or longer. The algorithm of the task, even if very simple, is not learned.

We further confirmed the overfitting problem on the machine translation task. This suggests that long-distance dependencies are not the only reason behind the decreased performance when translating very long sentences.
We think that our findings can shed a new light on specific areas of deep learning research, namely domain adaptation and curriculum learning.

We also showed that
data augmentation can tackle the data sparsity in the domain of very long sentences.

\section*{Acknowledgements}
This work was supported by
the GA ČR NEUREM3 grant (Neural Representations in Multi-modal and Multi-lingual Modelling, 19-26934X (RIV: GX19-26934X)) and by SVV 260 453 grant.

\bibliography{custom}
\bibliographystyle{acl_natbib}
\clearpage

\appendix
\section{Model Details}
\label{app:model}

In the following section, we provide the details of the used models and their training.
All the described models are implemented in Fairseq \citep{ott2019fairseq}.\footnote{\url{https://github.com/pytorch/fairseq}}
During training, we use word-level cross-entropy loss with teacher forcing \citep{bahdanau2015neural, vaswani2017attention} which is a current, widely used approach to the sequence-to-sequence Transformer training.
During decoding, we use beam search with beam size 4 and length penalty 0.6.

\subsection{String Editing}
\label{app:model-string}

In the experiments with string editing, we use the \texttt{transformer} parameter setting with the following modifications:
\begin{itemize}
    \item embeddings size: 128,
    \item feedforward size: 512,
    \item number of attention heads: 8,
    \item encoder/decoder depth: 1,
    \item batch size: 4,096 tokens,
    \item learning rate 5e-4,
    \item warmup steps: 4,000,
    \item dropout: 0.3,
    \item train epochs: 100
\end{itemize}

\subsection{Machine Translation}
\label{app:model-mt}

In the machine translation experiments, we use the \texttt{transformer} parameter setting with the following modifications:
\begin{itemize}
    \item embeddings size: 512,
    \item feedforward size: 2048,
    \item number of attention heads: 8,
    \item encoder/decoder depth: 6,
    \item batch size: 4,096,
    \item learning rate: 5e-4,
    \item warmup steps: 4,000,
    \item dropout: 0.3
\end{itemize}
During training, we apply early stopping: if the model performance in BLEU \cite{papineni2002bleu} does not improve for 10 epochs (evaluated on the complete held-out test set without length splits), the training is terminated.
\section{Translation Output Examples}
\label{app:ex}

Figure~\ref{fig:case-study-basic} shows example outputs from models trained on various target-length training buckets (10-, 30- and 60-bucket) produced by translating a chosen 30-bucket testset inputs.
The examples demonstrate that he models have tendencies to produce outputs with length similar to the training data while trying to satisfy the translation of the source sentence resulting in the longer, 60-bucket model repeating certain phrases or sentences while introducing grammatical errors (e.g. wrong agreement, preposition choice) or mistranslations.
On the other hand, the shorter, 10-bucket model manages to drop parts of the input sentence while maintaining a reasonable fluency and grammatical correctness of the output.

Figure~\ref{fig:case-study-concat} shows example of outputs from models trained on the synthetic 60-bucket data created by concatenation of the shorter training buckets.
At first glance, all three hypotheses are very similar and are reasonably good translations of the source sentence, however, all systems made a wrong surface form  and preposition choice for ``na Vinohradech'' (the same grammatical mistake as with ``na Žižkově'' in Figure~\ref{fig:case-study-basic}), producing an incorrect but literal translation of the English ``in Vinohrady''.
Additionally, all three systems chose a literal translation of the word ``approach'', which is incorrect in the given context.
The incorrect surface form of the translation ``založeno na doporučení'' in \textit{Hyp1} suggests that training a model on a concatenation of very short sentences may lead to incorrect modeling of long-range dependencies.
Surprisingly, the \textit{Hyp3} system mistranslated the phrase ``work on the reconstruction'' (``k rekonstrukci'' in the output) while \textit{Hyp2} system produced a correct translation, though this error is most likely a result of different set of training sentences in the \textit{Hyp2} and \textit{Hyp3} training data rather than the lenght of the training sentences (before concatenation).

\begin{figure*}[!ht]
\begin{center}
\small
\begin{tabularx}{\linewidth}{lX}
\toprule
\toprule
Source (30-bucket) & \makecell[Xt]{The company does not collect its mail and it has closed its official headquarters in Žižkov more than six years ago.} \\
\midrule

Hyp1 (10-bucket) & Společnost \blue{nesbírá poštu} a \underline{zavřel} \green{oficiální sídlo.} \\
\textit{Hyp1 (gloss)} & \textit{The company \blue{does not gather mail} and \underline{closed} \green{official headquarters.}} \\
\midrule

Hyp2 (30-bucket) & Společnost \red{neshromažďuje poštu} a \blue{již před více než šesti lety} zavřela své oficiální sídlo \green{\underline{v Žižkově}}. \\
\textit{Hyp2 (gloss)} & \textit{The company \red{does not collect mail} and \blue{more than six years ago} closed its official headquarters \green{\underline{in Žižkov}}.} \\
\midrule

Hyp3 (60-bucket) & \makecell[Xt]{Společnost \red{nevybírá poštu} a uzavřela \blue{své oficiální sídlo} \underline{v Žižkově} \green{více než šest let} \underline{ago}. v Žižkově. Společnost \violet{neshromažďuje poštu a uzavírá oficiální ústředí} \underline{v Žižkově více než šest let agr. o.}} \\
\textit{Hyp3 (gloss)} & \textit{The company \red{does not pick up mail} and closed up its official \blue{its official headquarters} \underline{in Žižkov} \green{more than six years} \underline{ago. in Žižkov}. The company \violet{does not collect mail and closes up official headquarters} \underline{in Žižkov more than six years agr. o.}} \\
\midrule

Reference (30-bucket) & \blue{Nepřebírá poštu} a oficiální sídlo na Žižkově \red{zrušila} před více než šesti lety. \\
\textit{Ref (gloss)} & \textit{(The company) \blue{does not collect mail} and official headquarters in Žižkov \red{closed up} more than six years ago.} \\

\midrule
\midrule

Source (30-bucket) & The perpetrators ended up in custody, said Marie Štrbáková, the spokeswoman of Olomouc police. \\
\midrule

Hyp1 (10-bucket) & \red{Mluvila s ní} Marie \underline{Štrkováková} \\
\textit{Hyp1 (gloss)} & \textit{\red{Talked to her}, Marie \underline{Štrkováková}} \\
\midrule

Hyp2 (30-bucket) & Pachatelé \blue{skončili ve vazbě}, řekla Marie \underline{Štrbákováová}, \green{mluvčí Olomouckého policie.} \\
\textit{Hyp2 (gloss)} & \textit{The perpetrators \blue{ended up in custody}, said Marie \underline{Štrbákováová}, \green{the spokeswoman of Olomouc police.}} \\
\midrule

Hyp3 (60-bucket) & \makecell[Xt]{\underline{Uchazeči} \red{skoncovali} \underline{v úschově}, \blue{\underline{"uvedla} Marie \underline{Štrbákováová},} mluvčí Olomoucké policie\underline{\green{, která se stala mluvčí} Olomouckého vojska, \violet{a to v úschově.}}} \\
\textit{Hyp3 (gloss)} & \textit{\underline{The candidates} \red{ended up} \underline{in storage}, \blue{\underline{"introduced} Marie \underline{Štrbákováová},} the spokeswoman of Olomouc police\underline{\green{, which became the spokeswoman} of Olomouc army, \violet{and in storage.}}} \\
\midrule

Ref (30-bucket) & Pachatelé \blue{skončili ve vazbě}, informovala olomoucká policejní \green{mluvčí Marie Štrbáková.} \\
\textit{Ref (gloss)} & \textit{The perpetrators \blue{ended up in custody}, informed Olomouc police \green{spokeswoman Marie Štrbáková.}} \\
\bottomrule
\bottomrule
\end{tabularx}
\end{center}
\caption{Example translations from systems trained on specific target-length-restricted datasets. Both examples demonstrate over and under-generation of systems trained on datasets containing longer (60-bucket) and shorter (10-bucket) sentences when applied to inputs with length of reference translation different from the training data (30-bucket). We provide rough, ``word-for-word'' translations of the produced outputs (in \textit{italics}) with color highlighting of some of the phrases and their corresponding English translation for better comprehension. The \underline{underline} highlights grammatical errors or mistranslations in the output.}
\label{fig:case-study-basic}
\end{figure*}

\begin{figure*}[!ht]
\begin{center}
\small
\begin{tabularx}{\linewidth}{lX}
\toprule
\toprule
Source (60-bucket) & \makecell[Xt]{\red{We have already worked with} Lenka Langerová \blue{on our flat in the mountains} based on a recommendation from another client \green{and because everything worked well} we decided to approach her \violet{to work on the reconstruction} of our new flat in Vinohrady.} \\
\midrule

Hyp1 (10-bucket-concat) & \makecell[Xt]{\red{Už jsme pracovali s} Lenkou Langerovou \blue{na našem bytě v horách} \underline{založeno na doporučení} od jiného klienta \green{a protože všechno fungovalo dobře,} rozhodli jsme se k ní \underline{přiblížit} \violet{\underline{k práci} na rekonstrukci} našeho nového bytu \underline{ve Vinohrady.}} \\
\midrule

Hyp2 (30-bucket-concat) & \makecell[Xt]{\red{Již jsme spolupracovali s} Lenkou Langerovou \blue{na našem bytě v horách} na základě doporučení jiného klienta \green{a protože vše fungovalo dobře,} rozhodli jsme se, že se k ní \underline{přiblížíme}\violet{, aby pracovala na rekonstrukci} našeho nového bytu \underline{ve Vinohrady.}} \\
\midrule

Hyp3 (60-bucket) & \makecell[Xt]{\red{Již jsme s} Lenkou Langerovou \red{spolupracovali} \blue{na našem \underline{bytu} v horách} na základě doporučení jiného  klienta \green{a protože vše fungovalo dobře,} rozhodli jsme se, že se k ní \underline{přiblížíme} \violet{\underline{k rekonstrukci}} našeho nového bytu \underline{ve Vinohrady.}} \\
\midrule

Ref (60-bucket) & \makecell[Xt]{\red{S} architektkou Lenkou Langerovou \red{jsme spolupracovali už} \blue{na našem horském apartmánu}, tehdy na bázi osobního doporučení jiného klienta, \green{a vzhledem k tomu, že vše dobře fungovalo,} byla pro nás jasná volba \violet{i při rekonstrukci} našeho nového bytu na Vinohradech.} \\
\bottomrule
\bottomrule
\end{tabularx}
\end{center}
\caption{Example of translation hypotheses generated by a system trained on a genuine 60-bucket data and systems trained only on a concatenation of shorter training examples (10-bucket-concat, 30-bucket-concat) for comparison. Color highlighting indicates the correspondence of Czech and English phrases. The \underline{underline} highlights grammatical errors in the output.}
\label{fig:case-study-concat}
\end{figure*}
\section{Source-Side Bucketing Experiments}
\label{app:src-experiments}

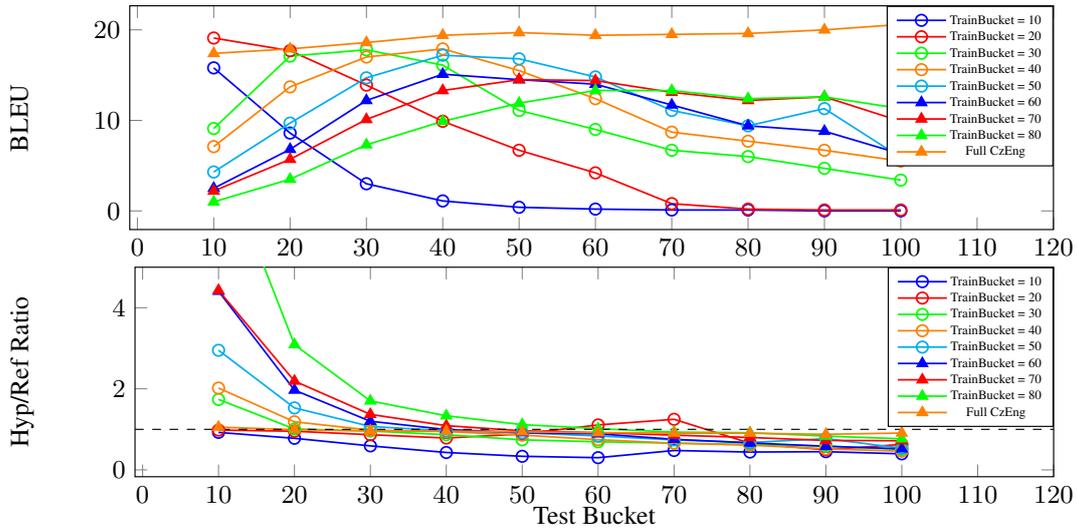
\begin{figure*}[t]
    \centering
    \begin{minipage}[b]{\linewidth}
        \centering
        \def\verticalstep{0.8}
\small
\pgfplotscreateplotcyclelist{buckets}{
semithick,blue,mark=o\\%
semithick,red,mark=o\\%
semithick,green,mark=o\\%
semithick,orange,mark=o\\%
semithick,cyan,mark=o\\%
semithick,blue,mark=triangle*\\%
semithick,red,mark=triangle*\\%
semithick,green,mark=triangle*\\%
semithick,orange,mark=triangle*\\%
semithick,cyan,mark=triangle*\\%
}

\begin{tikzpicture}[]
  \begin{axis}[cycle list name=buckets, 
               xlabel={},
               ylabel={BLEU},
               xmax=120,
               width=0.8\textwidth,
               height=0.27\textwidth,
               x label style={at={(axis description cs:0.5,-0.1)},anchor=north},
               y label style={at={(axis description cs:-0.1,.5)},rotate=0,anchor=south},
               legend style={nodes={scale=0.5, transform shape},at={(1,1)},anchor=north east}]

    \addplot+ table [x=BUCKET, y=BLEU, col sep=comma] {./data/buckets/buckets.src.10.csv};
    \addplot+ table [x=BUCKET, y=BLEU, col sep=comma] {./data/buckets/buckets.src.20.csv};
    \addplot+ table [x=BUCKET, y=BLEU, col sep=comma] {./data/buckets/buckets.src.30.csv};
    \addplot+ table [x=BUCKET, y=BLEU, col sep=comma] {./data/buckets/buckets.src.40.csv};
    \addplot+ table [x=BUCKET, y=BLEU, col sep=comma] {./data/buckets/buckets.src.50.csv};
    \addplot+ table [x=BUCKET, y=BLEU, col sep=comma] {./data/buckets/buckets.src.60.csv};
    \addplot+ table [x=BUCKET, y=BLEU, col sep=comma] {./data/buckets/buckets.src.70.csv};
    \addplot+ table [x=BUCKET, y=BLEU, col sep=comma] {./data/buckets/buckets.src.80.csv};
    \addplot+ table [x=BUCKET, y=BLEU, col sep=comma] {./data/buckets/buckets.csv};
    \addlegendentry{TrainBucket = 10}
    \addlegendentry{TrainBucket = 20}
    \addlegendentry{TrainBucket = 30}
    \addlegendentry{TrainBucket = 40}
    \addlegendentry{TrainBucket = 50}
    \addlegendentry{TrainBucket = 60}
    \addlegendentry{TrainBucket = 70}
    \addlegendentry{TrainBucket = 80}
    \addlegendentry{Full CzEng}
  \end{axis}
\end{tikzpicture}
        \def\verticalstep{0.8}
\small
\pgfplotscreateplotcyclelist{buckets}{
semithick,blue,mark=o\\%
semithick,red,mark=o\\%
semithick,green,mark=o\\%
semithick,orange,mark=o\\%
semithick,cyan,mark=o\\%
semithick,blue,mark=triangle*\\%
semithick,red,mark=triangle*\\%
semithick,green,mark=triangle*\\%
semithick,orange,mark=triangle*\\%
semithick,cyan,mark=triangle*\\%
}

\begin{tikzpicture}[]
  \begin{axis}[cycle list name=buckets, 
               xlabel={Test Bucket},
               ylabel={Hyp/Ref Ratio},
               xmax=120,
               ymax=5,
               width=0.8\textwidth,
               height=0.26\textwidth,
               x label style={at={(axis description cs:0.5,-0.1)},anchor=north},
               y label style={at={(axis description cs:-0.1,.5)},rotate=0,anchor=south},
               legend style={nodes={scale=0.5, transform shape},at={(1,1)},anchor=north east}]

    \addplot+ table [x=BUCKET, y=RATIO, col sep=comma] {./data/ratios/ratios.src.10.csv};
    \addplot+ table [x=BUCKET, y=RATIO, col sep=comma] {./data/ratios/ratios.src.20.csv};
    \addplot+ table [x=BUCKET, y=RATIO, col sep=comma] {./data/ratios/ratios.src.30.csv};
    \addplot+ table [x=BUCKET, y=RATIO, col sep=comma] {./data/ratios/ratios.src.40.csv};
    \addplot+ table [x=BUCKET, y=RATIO, col sep=comma] {./data/ratios/ratios.src.50.csv};
    \addplot+ table [x=BUCKET, y=RATIO, col sep=comma] {./data/ratios/ratios.src.60.csv};
    \addplot+ table [x=BUCKET, y=RATIO, col sep=comma] {./data/ratios/ratios.src.70.csv};
    \addplot+ table [x=BUCKET, y=RATIO, col sep=comma] {./data/ratios/ratios.src.80.csv};
    \addplot+ table [x=BUCKET, y=RATIO, col sep=comma] {./data/ratios/ratios.csv};
    
    \coordinate (A) at (0,1.0);
    \coordinate (O1) at (rel axis cs:0,0);
    \coordinate (O2) at (rel axis cs:1,0);
    \draw [black,sharp plot,dashed] (A -| O1) -- (A -| O2);
    
    \addlegendentry{TrainBucket = 10}
    \addlegendentry{TrainBucket = 20}
    \addlegendentry{TrainBucket = 30}
    \addlegendentry{TrainBucket = 40}
    \addlegendentry{TrainBucket = 50}
    \addlegendentry{TrainBucket = 60}
    \addlegendentry{TrainBucket = 70}
    \addlegendentry{TrainBucket = 80}
    \addlegendentry{Full CzEng}
  \end{axis}
\end{tikzpicture}
    \end{minipage} 
    \caption{\textbf{Top}: Varying performance of Transformers on test data trained only on the data from a specific source-side length bucket (various lines) when evaluated on a specific test bucket (x-axis). BLEU scores are not directly comparable across different test sets (i.e. horizontally). \textbf{Bottom}: Average ratio between a hypothesis and reference. Dashed line indicates a ratio of 1.0.}
    \label{fig:mt-length-src}
\end{figure*}

For comparison, we repeated the translation experiments using source-side length-based bucketing of the training and validation data.
Figure~\ref{fig:mt-length-src} shows the performance of the bucketed models with respect to testset of various bucket sizes.
While the results are similar to the target-side bucketing experiments, the overfitting phenomenon is less clear in several cases (e.g. 20-bucket system reaching higher BLEU than 10-bucket system on 10-bucket testset or the relative system ranking on the 60-bucket testset).

\begin{figure}[t]
    \centering
    \begin{minipage}[b]{\linewidth}
        \centering
        \begin{tikzpicture}
    \begin{axis}
        [
        xmin=0,
        xmax=300,
        xlabel={Target Length},
        ylabel={Source Length},
        ytick={1,2,3,4,5,6,7,8},
        yticklabels={0-10, 11-20, 21-30, 31-40, 41-50, 51-60, 61-70, 71-80},
        ]
\addplot+[
    boxplot prepared={
        median=6.0,
        upper quartile=8.0,
        lower quartile=4.0,
        upper whisker=111.0,
        lower whisker=1.0
    },
] coordinates {};
\addplot+[
    boxplot prepared={
        median=14.0,
        upper quartile=17.0,
        lower quartile=12.0,
        upper whisker=167.0,
        lower whisker=1.0
    },
] coordinates {};
\addplot+[
    boxplot prepared={
        median=25.0,
        upper quartile=28.0,
        lower quartile=21.0,
        upper whisker=209.0,
        lower whisker=2.0
    },
] coordinates {};
\addplot+[
    boxplot prepared={
        median=35.0,
        upper quartile=39.0,
        lower quartile=31.0,
        upper whisker=192.0,
        lower whisker=3.0
    },
] coordinates {};
\addplot+[
    boxplot prepared={
        median=45.0,
        upper quartile=49.0,
        lower quartile=40.0,
        upper whisker=281.0,
        lower whisker=4.0
    },
] coordinates {};
\addplot+[
    boxplot prepared={
        median=54.0,
        upper quartile=60.0,
        lower quartile=49.0,
        upper whisker=227.0,
        lower whisker=4.0
    },
] coordinates {};
\addplot+[
    boxplot prepared={
        median=64.0,
        upper quartile=70.0,
        lower quartile=58.0,
        upper whisker=339.0,
        lower whisker=2.0
    },
] coordinates {};
\addplot+[
    boxplot prepared={
        median=74.0,
        upper quartile=81.0,
        lower quartile=67.0,
        upper whisker=283.0,
        lower whisker=4.0
    },
] coordinates {};
 
    \end{axis}
\end{tikzpicture}
        \begin{tikzpicture}
    \begin{axis}
        [
        xmin=0,
        xmax=300,
        xlabel={Target Length},
        ylabel={Source Length},
        ytick={1,2,3,4,5,6,7,8},
        yticklabels={0-10, 11-20, 21-30, 31-40, 41-50, 51-60, 61-70, 71-80},
        ]
\addplot+[
    boxplot prepared={
        median=5.0,
        upper quartile=6.0,
        lower quartile=4.0,
        upper whisker=17.0,
        lower whisker=1.0
    },
] coordinates {};
\addplot+[
    boxplot prepared={
        median=10.0,
        upper quartile=12.0,
        lower quartile=8.0,
        upper whisker=23.0,
        lower whisker=2.0
    },
] coordinates {};
\addplot+[
    boxplot prepared={
        median=16.0,
        upper quartile=18.0,
        lower quartile=13.0,
        upper whisker=33.0,
        lower whisker=6.0
    },
] coordinates {};
\addplot+[
    boxplot prepared={
        median=22.0,
        upper quartile=25.0,
        lower quartile=19.0,
        upper whisker=39.0,
        lower whisker=11.0
    },
] coordinates {};
\addplot+[
    boxplot prepared={
        median=28.0,
        upper quartile=31.0,
        lower quartile=25.0,
        upper whisker=46.0,
        lower whisker=10.0
    },
] coordinates {};
\addplot+[
    boxplot prepared={
        median=34.0,
        upper quartile=38.0,
        lower quartile=31.0,
        upper whisker=53.0,
        lower whisker=19.0
    },
] coordinates {};
\addplot+[
    boxplot prepared={
        median=40.0,
        upper quartile=46.0,
        lower quartile=36.0,
        upper whisker=58.0,
        lower whisker=24.0
    },
] coordinates {};
\addplot+[
    boxplot prepared={
        median=47.0,
        upper quartile=51.0,
        lower quartile=42.0,
        upper whisker=66.0,
        lower whisker=26.0
    },
] coordinates {};        
        
    \end{axis}
\end{tikzpicture}
    \end{minipage} 
    \caption{Distribution of lengths of target-side references within the training (\textbf{top}) and validation (\textbf{bottom}) datasets after splitting them into source-side length buckets.
    Both figures have identical x-axis scaling for better comparison. The long whiskers of the training bucket length distributions are a result of a noisy nature of CzEng 2.0 training corpus.
    }
    \label{fig:tgt-length-distro}
\end{figure}
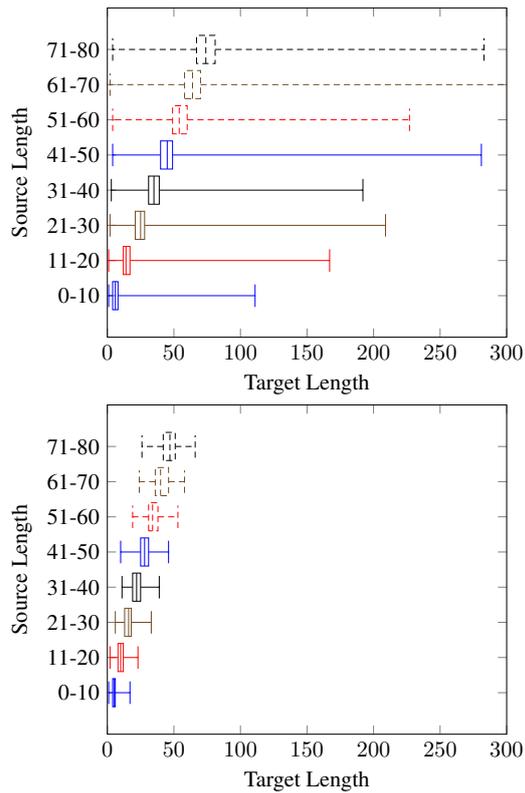

We think that the possible reason is the difference between the source-side length and the length of training/validation reference leading to possible overlap of target-side lengths between the different train/validation buckets.
Figure~\ref{fig:tgt-length-distro} shows the length distributions of target-side lengths within each training and validation bucket.
Although the length-wise overlap between the target-side of the training/validation examples is manifested mostly in the $1^{st}$ and $4^{th}$ quartile, we think that it helps to support the argument that the length-based overfitting should be studied with respect to the target-side length instead of the source-side.
Furthermore, the length of the target-side (Czech) in the test dataset is generally smaller than the source-side (English), resulting in additional domain mismatch between the training-test buckets.
Note that very long target-side outliers in the training data are most likely a result of an imperfect sentence-pair filtering after the inclusion of the additional synthetic parallel data (forward and backward translation) to the CzEng 2.0 corpus.

\begin{figure*}[t]
    \centering
    \begin{minipage}[b]{\linewidth}
        \centering
        \def\verticalstep{0.8}
\small
\pgfplotscreateplotcyclelist{buckets}{
semithick,blue,mark=o\\%
semithick,red,mark=o\\%
semithick,green,mark=o\\%
semithick,orange,mark=o\\%
semithick,cyan,mark=o\\%
semithick,blue,mark=triangle*\\%
semithick,red,mark=triangle*\\%
semithick,green,mark=triangle*\\%
semithick,orange,mark=triangle*\\%
semithick,cyan,mark=triangle*\\%
}

\begin{tikzpicture}[]
  \begin{axis}[cycle list name=buckets, 
               xlabel={},
               ylabel={BLEU},
               xmax=120,
               width=0.8\textwidth,
               height=0.27\textwidth,
               x label style={at={(axis description cs:0.5,-0.1)},anchor=north},
               y label style={at={(axis description cs:-0.1,.5)},rotate=0,anchor=south},
               legend style={nodes={scale=0.5, transform shape},at={(1,1)},anchor=north east}]

    \addplot+ table [x=BUCKET, y=BLEU, col sep=comma] {./data/buckets/buckets.src.tuning.10.csv};
    \addplot+ table [x=BUCKET, y=BLEU, col sep=comma] {./data/buckets/buckets.src.tuning.20.csv};
    \addplot+ table [x=BUCKET, y=BLEU, col sep=comma] {./data/buckets/buckets.src.tuning.30.csv};
    \addplot+ table [x=BUCKET, y=BLEU, col sep=comma] {./data/buckets/buckets.src.tuning.40.csv};
    \addplot+ table [x=BUCKET, y=BLEU, col sep=comma] {./data/buckets/buckets.src.tuning.50.csv};
    \addplot+ table [x=BUCKET, y=BLEU, col sep=comma] {./data/buckets/buckets.src.tuning.60.csv};
    \addplot+ table [x=BUCKET, y=BLEU, col sep=comma] {./data/buckets/buckets.src.tuning.70.csv};
    \addplot+ table [x=BUCKET, y=BLEU, col sep=comma] {./data/buckets/buckets.src.tuning.80.csv};
    \addplot+ table [x=BUCKET, y=BLEU, col sep=comma] {./data/buckets/buckets.csv};
    \addlegendentry{TrainBucket = 10}
    \addlegendentry{TrainBucket = 20}
    \addlegendentry{TrainBucket = 30}
    \addlegendentry{TrainBucket = 40}
    \addlegendentry{TrainBucket = 50}
    \addlegendentry{TrainBucket = 60}
    \addlegendentry{TrainBucket = 70}
    \addlegendentry{TrainBucket = 80}
    \addlegendentry{Full CzEng}
  \end{axis}
\end{tikzpicture}
        \def\verticalstep{0.8}
\small
\pgfplotscreateplotcyclelist{buckets}{
semithick,blue,mark=o\\%
semithick,red,mark=o\\%
semithick,green,mark=o\\%
semithick,orange,mark=o\\%
semithick,cyan,mark=o\\%
semithick,blue,mark=triangle*\\%
semithick,red,mark=triangle*\\%
semithick,green,mark=triangle*\\%
semithick,orange,mark=triangle*\\%
semithick,cyan,mark=triangle*\\%
}

\begin{tikzpicture}[]
  \begin{axis}[cycle list name=buckets, 
               xlabel={Test Bucket},
               ylabel={Hyp/Ref Ratio},
               xmax=120,
               ymax=5,
               width=0.8\textwidth,
               height=0.26\textwidth,
               x label style={at={(axis description cs:0.5,-0.1)},anchor=north},
               y label style={at={(axis description cs:-0.1,.5)},rotate=0,anchor=south},
               legend style={nodes={scale=0.5, transform shape},at={(1,1)},anchor=north east}]

    \addplot+ table [x=BUCKET, y=RATIO, col sep=comma] {./data/ratios/ratios.src.tuning.10.csv};
    \addplot+ table [x=BUCKET, y=RATIO, col sep=comma] {./data/ratios/ratios.src.tuning.20.csv};
    \addplot+ table [x=BUCKET, y=RATIO, col sep=comma] {./data/ratios/ratios.src.tuning.30.csv};
    \addplot+ table [x=BUCKET, y=RATIO, col sep=comma] {./data/ratios/ratios.src.tuning.40.csv};
    \addplot+ table [x=BUCKET, y=RATIO, col sep=comma] {./data/ratios/ratios.src.tuning.50.csv};
    \addplot+ table [x=BUCKET, y=RATIO, col sep=comma] {./data/ratios/ratios.src.tuning.60.csv};
    \addplot+ table [x=BUCKET, y=RATIO, col sep=comma] {./data/ratios/ratios.src.tuning.70.csv};
    \addplot+ table [x=BUCKET, y=RATIO, col sep=comma] {./data/ratios/ratios.src.tuning.80.csv};
    \addplot+ table [x=BUCKET, y=RATIO, col sep=comma] {./data/ratios/ratios.csv};
    
    \coordinate (A) at (0,1.0);
    \coordinate (O1) at (rel axis cs:0,0);
    \coordinate (O2) at (rel axis cs:1,0);
    \draw [black,sharp plot,dashed] (A -| O1) -- (A -| O2);
    
    \addlegendentry{TrainBucket = 10}
    \addlegendentry{TrainBucket = 20}
    \addlegendentry{TrainBucket = 30}
    \addlegendentry{TrainBucket = 40}
    \addlegendentry{TrainBucket = 50}
    \addlegendentry{TrainBucket = 60}
    \addlegendentry{TrainBucket = 70}
    \addlegendentry{TrainBucket = 80}
    \addlegendentry{Full CzEng}
  \end{axis}
\end{tikzpicture}
    \end{minipage} 
    \caption{\textbf{Top}: Varying performance of Transformers on test data trained on all of CzEng and fine-tuned only on the data from a specific source-side length bucket (various lines) when evaluated on a specific test bucket (x-axis). BLEU scores are not directly comparable across different test sets (i.e. horizontally). \textbf{Bottom}: Average ratio between a hypothesis and reference. Dashed line indicates a ratio of 1.0. We preserve the scaling of all the plots for better comparability across the figures.
    }
    \label{fig:mt-length-src-tuning}
\end{figure*}
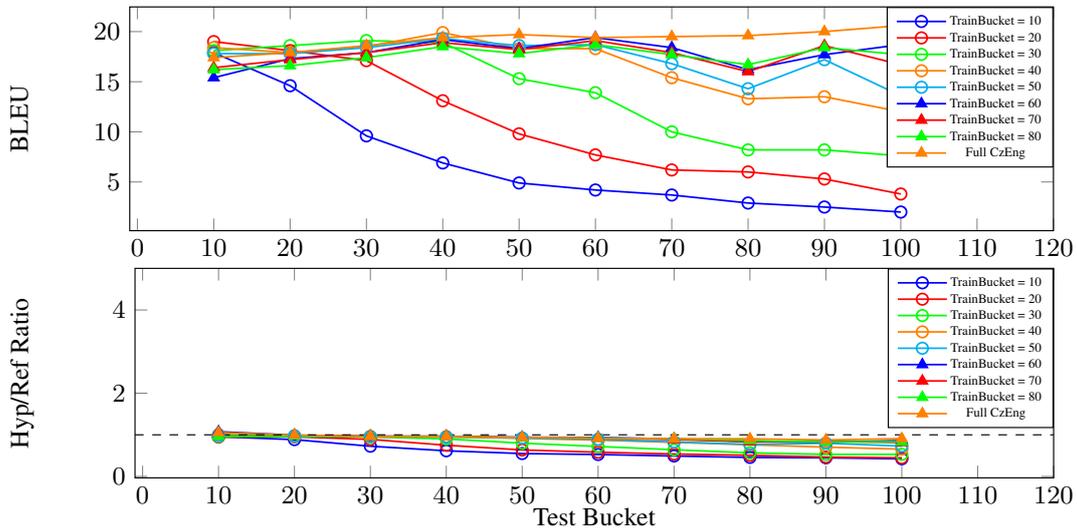

Based on the reviewer's suggestion, we also measured the effect of finetuning a system trained on the whole training dataset using a source-side bucketed training data.
Each system was fine-tuned for 30 epochs, although, it is important to note that the validation BLEU of each fine-tuned system was dropping during training (compared to the BLEU of the initial model) when evaluated against the whole non-bucketed validation dataset.
In Figure~\ref{fig:mt-length-src-tuning}, we can see a growing effect of catastrophic forgetting \citep{kirkpatrick2017catastrophic}: all models initially saw all lengths during pretraining but specialized for a specific length bucket during finetuning.
Interestingly, the forgetting effect is stronger for test buckets that are longer than the finetuning lengths while the models show much better retention of the ability to model shorter sentences.

Lastly, we also performed a comparison between the baseline MT system and the combination of systems trained on a specific source-side length buckets training datasets.
We extracted sentences from our test dataset with source-side length 0--80, translated them with the respective systems and computed the BLEU scores using MultEval \citep{clark2011multeval}.\footnote{\url{https://github.com/jhclark/multeval}}
We compared a system combination trained using only a specific length-bucket dataset (\texttt{bucketed}) applied on the respective ``in-domain'' parts of the test dataset.
We also provide comparison with the system combination initialized by the baseline model and then fine-tuned on the respective length-bucket datasets (\texttt{bucketed.tuning}).
Additionally, we also trained a system using CzEng 2.0 with additional source-side labels indicating a length-bucket in which a given training example ended up after the source-side length-based dataset splitting (\texttt{bucket.labels}), e.g ``<20> Example sentence...'' for a sentence from a bucket 11--20.
This model was evaluated on the same test dataset with inclusion of these source-side length-bucket labels.

\begin{table}[t]
\begin{center}
\small
\begin{tabular}{r|r}
\toprule
 & BLEU \\
\midrule
baseline & 19.1 $\pm$ 0.2\\
\midrule
bucketed & 18.9 $\pm$ 0.2 \\
bucketed.tuning & 17.1 $\pm$ 0.2 \\
bucket.labels & 16.7 $\pm$ 0.2 \\
\bottomrule
\end{tabular}
\caption{Comparison of the translation performance of the  \texttt{baseline} model trained on the whole CzEng 2.0, and source-length specialized models. \texttt{bucketed} is a combination of systems trained on the source-length bucketed training data, \texttt{bucketed.tuning} is a similar combination, where systems were first initialized by the \texttt{baseline} model and then fine-tuned for 30 epochs on their respective buckets. \texttt{bucket.labels} is a system trained on the whole CzEng 2.0 with inclusion of the source-side bucket length labels on the input. The systems were evaluated using MultEval \citep{clark2011multeval} using a bootstrapping over a test dataset containing sentences of source-side lengths 0--80. Only a single optimizer run was performed for each evaluated system.} 
\label{tab:sys-comparison}
\end{center}
\end{table}

The results in Table~\ref{tab:sys-comparison} suggest that the length-based specialization of the models does not outperform the baseline.
One of the possible explanations is a fact that \texttt{baseline} system was trained on the whole CzEng 2.0 containing even sentences longer than 80.
Although the \texttt{bucket.labels} was also trained using the whole CzEng 2.0, the results suggest that a simple inclusion of the source-length bucket information does not contribute towards a better translation performance.


\newwrite\outputstream
\immediate\openout\outputstream=appendix.tmp
\immediate\write\outputstream{\thepage}
\immediate\closeout\outputstream

\end{document}